\def\cvprPaperID{7319}
\def\confName{CVPR}
\def\confYear{2023}

\def\paperTitle{\ours{}: Hallucinating Latent Positives for Skeleton-based Self-Supervised Learning of Actions}

\def\authorBlock{
    Anshul Shah$^1$ \qquad
    Aniket Roy$^1$\thanks{Aniket Roy and Ketul Shah contributed equally} \qquad Ketul Shah$^1$\footnotemark[1] \qquad Shlok Mishra$^2$ \qquad \\
    David Jacobs$^{2,3}$ \qquad Anoop Cherian$^4$ \qquad Rama Chellappa$^1$ \\
    $^1$Johns Hopkins University \qquad $^2$University of Maryland, College Park \qquad $^3$Meta \qquad $^4$MERL\\
    {\tt\small \{ashah95, aroy28, kshah33, rchella4\}@jhu.edu \quad  \{shlokm, dwj\}@umd.edu \quad cherian@merl.com}
}

\newif\ifreview 
\newif\ifarxiv \newcommand{\arxiv}{\arxivtrue}
\newif\ifcamera 
\newif\ifrebuttal 

\newcommand{\inpske}{$x$\xspace}
\newcommand{\encoder}{$E$\xspace}

\newcommand{\xk}{$x_k$\xspace}
\newcommand{\xq}{$x_q$\xspace}
    \newcommand{\encoderq}{$E_q$\xspace}
\newcommand{\encoderk}{$E_k$\xspace}
\newcommand{\queue}{$Q$\xspace}
\newcommand{\zk}{$z_k$\xspace}
\newcommand{\zq}{$z_q$\xspace}
\newcommand{\zg}{$z^H_i$\xspace}

\newcommand{\direction}{$d$\xspace}
\newcommand{\selectedproto}{${P}_{\text{sel}}$\xspace}
\newcommand{\closestproto}{$P^*_{z_k}$\xspace}
\newcommand{\protoset}{$\mathcal{P}$\xspace}

\arxiv %

\pdfoutput=1
\documentclass[10pt,twocolumn,letterpaper]{article}
\ifreview \usepackage[review]{cvpr} \fi
\ifarxiv \usepackage[pagenumbers]{cvpr} \fi
\ifrebuttal \usepackage[rebuttal]{cvpr} \fi
\ifcamera \usepackage{cvpr} \fi

\usepackage{graphicx}
\usepackage{amsmath}
\usepackage{amssymb}
\usepackage{booktabs}

\usepackage{times}
\usepackage{microtype}
\usepackage{epsfig}
\usepackage[table,xcdraw]{xcolor}
\usepackage{caption}
\usepackage{float}
\usepackage{placeins}
\usepackage{color, colortbl}
\usepackage{stfloats}
\usepackage{enumitem}
\usepackage{tabularx}
\usepackage{xstring}
\usepackage{multirow}
\usepackage{xspace}
\usepackage{url}
\usepackage{subcaption}
\usepackage{xcolor}
\usepackage[hang,flushmargin]{footmisc}
\usepackage{multirow}
\usepackage{algorithm}
\usepackage{algorithmic}
\usepackage{enumitem,amssymb}
\usepackage{lipsum}
\ifcamera \usepackage[accsupp]{axessibility} \fi

\ifarxiv  \fi

\usepackage{pifont}

\usepackage{lipsum}

\newcommand{\ours}{HaLP}  %

\usepackage{xr-hyper}

\makeatletter
\newcommand*{\addFileDependency}[1]{
  \typeout{(#1)}
  \@addtofilelist{#1}
  \IfFileExists{#1}{}{\typeout{No file #1.}}
}

\makeatother

\usepackage[pagebackref,breaklinks,colorlinks]{hyperref}
\usepackage[capitalize]{cleveref}
\crefname{section}{Sec.}{Secs.}
\crefname{table}{Table}{Tables}
\crefname{figure}{Fig.}{Figs.}

\begin{document}
\title{\paperTitle}
\author{\authorBlock}
\maketitle

\begin{abstract}
Supervised learning of skeleton sequence encoders for action recognition has received significant attention in recent times. However, learning such encoders without labels continues to be a challenging problem. While prior works have shown promising results by applying contrastive learning to pose sequences, the quality of the learned representations is often observed to be closely tied to data augmentations that are used to craft the positives. However, augmenting pose sequences is a difficult task as the geometric constraints among the skeleton joints need to be enforced to make the augmentations realistic for that action. In this work, we propose a new contrastive learning approach to train models for skeleton-based action recognition without labels. Our key contribution is a simple module, \emph{HaLP} -- to Hallucinate Latent Positives for contrastive learning. Specifically, HaLP explores the latent space of poses in suitable directions to generate new positives. To this end, we present a novel optimization formulation to solve for the synthetic positives with an explicit control on their hardness. We propose approximations to the objective, making them solvable in closed form with minimal overhead. We show via experiments that using these generated positives within a standard contrastive learning framework leads to consistent improvements across benchmarks such as NTU-60, NTU-120, and PKU-II on tasks like linear evaluation, transfer learning, and kNN evaluation. Our code will be made available at \href{https://github.com/anshulbshah/HaLP}{https://github.com/anshulbshah/HaLP}.
\end{abstract}
\section{Introduction}
\label{sec:intro}
\begin{figure}[tp]
    \centering
    \includegraphics[trim={2cm 2cm 2cm 2cm},clip,width=\linewidth]{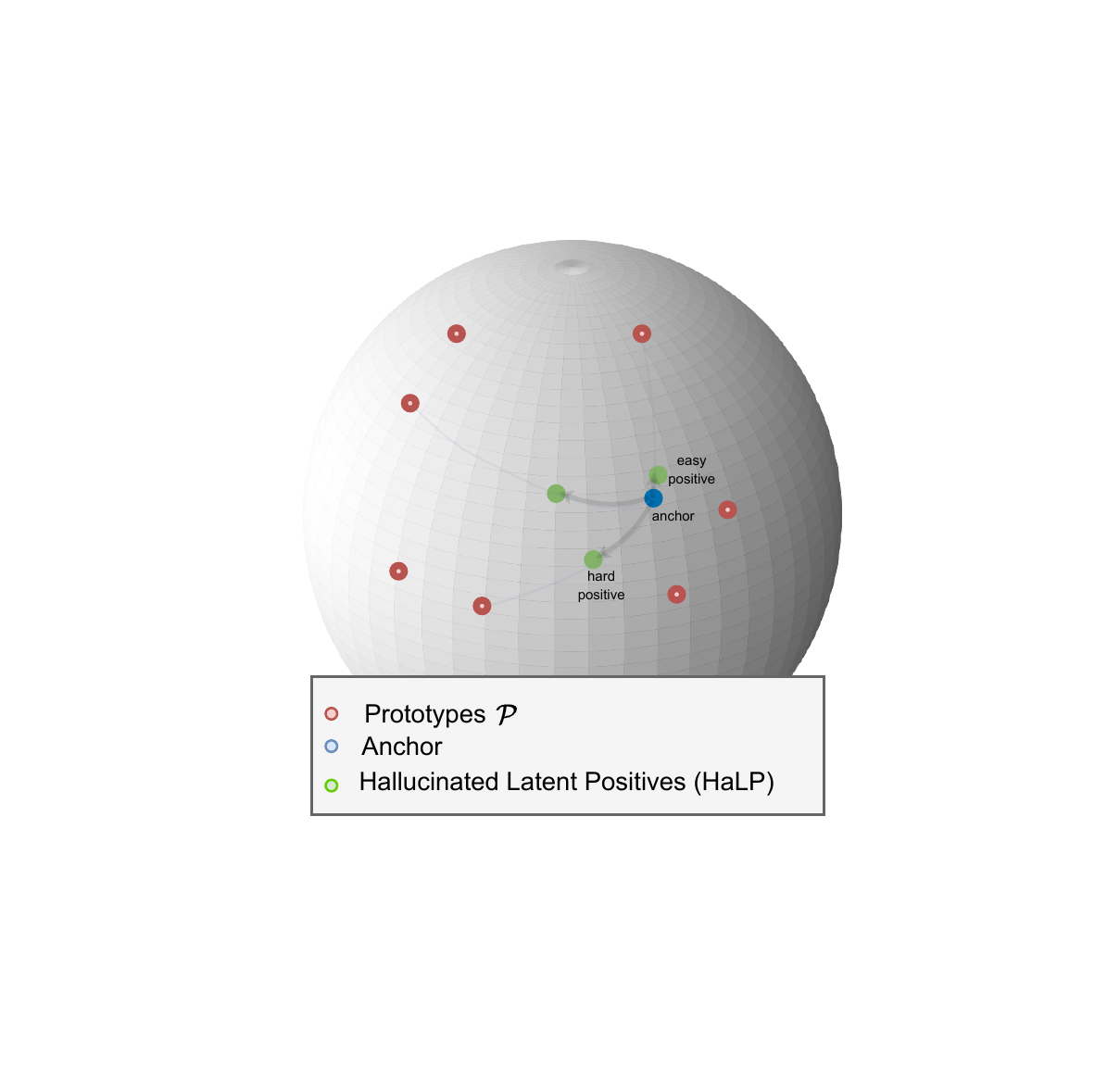}
    \caption{\ours{}: We propose an approach to hallucinate latent positives for use within a contrastive learning pipeline. Our approach works as follows: 1) We extract prototypes which succinctly represent the data at a particular step in training, 2) We randomly select a prototype from the prototype set, 3) Our approach then determines an optimal vector which when added to the anchor can generate positives of varying hardness. We generate a number of positives using this approach which are then used within a contrastive learning pipeline to train a model without labels.}
    \label{fig:teaser_figure}
\end{figure}

Recognizing human actions from videos is of immense practical importance with applications in behavior understanding~\cite{wei2014skeleton}, medical assistive applications~\cite{chang2011kinect}, AR/VR applications~\cite{kazakos2019epic} and surveillance~\cite{corona2021meva}. Action recognition has been an active area of research~\cite{kong2022human} with a focus on temporal understanding~\cite{feichtenhofer2019slowfast}, faster and efficient models for understanding complex actions~\cite{meng2021adafuse}, etc. Most works in the past have focused on action recognition from appearance information. But recent methods have shown the advantages of using pose/skeleton information as a separate cue with benefits in robustness to scene and object biases~\cite{li2018resound,li2019repair,weinzaepfel2021mimetics}, reduced privacy concerns~\cite{kidzinski2020deep}, apart from succinctly representing human motion~\cite{johansson1973visual}. However, annotating videos for skeleton-based action recognition is an arduous task and is difficult to scale. Prior work in self-supervised learning has shown advantages of learning without labels - including improved transfer performance \cite{chen2020simple, he2020momentum}, robustness to domain shift \cite{ge2021robust, pmlr-v162-wang22q} and noisy labels \cite{chuang2020debiased, Ghosh2021ContrastiveLI}. Inspired by these methodologies, we propose a new approach for skeleton-based action recognition without using labels.  

There have been several interesting approaches to tackling self-supervision for skeleton sequences. Methods like \cite{zhang2016colorful,noroozi2016unsupervised} have proposed improved pretext tasks to train models. Image-based self-supervised learning has shown impressive success using contrastive learning (CL)-based losses. Inspired by these, some recent approaches~\cite{thoker2021skeleton,mao2022cmd} have successfully applied CL to skeleton sequences, with modifications such as data augmentations~\cite{thoker2021skeleton}, use of multi-modal cues~\cite{mao2022cmd}, etc. The success of CL for a problem is closely tied to the data augmentations used to create the positives and the quality and number of negatives used to offer contrast to the positives~\cite{chen2020simple,he2020momentum}. While various works have tried focusing on negatives for improving the performance of Skeleton-SSL models, augmenting skeleton sequences is more difficult. Unlike images, skeletons are geometric structures, and devising novel data augmentations to craft new positives is an interesting but difficult task. 

In this work, we address the question of whether we can hallucinate new positives in the latent space~(\cref{fig:teaser_figure}) for use in a CL framework. Our approach, which we call \ours{}: \textbf{Ha}llucinating \textbf{L}atent \textbf{P}ositives has dual benefits; generating positives in latent space can reduce reliance on hand-crafted data augmentations. Further, it allows for faster training than using multiple-view approaches~\cite{caron2020unsupervised,caron2021emerging}, which incur significant overheads. Recall that, CL trains a model by pulling two augmented versions of a skeleton sequence close in the latent space while pushing them far apart from the negatives. Our key idea in this work is to hallucinate new positives, thus exploring new parts of the latent space beyond the query and key to improve the learning process. We introduce two new components to the CL pipeline. The first extracts prototypes from the data which succinctly represent the high dimensional latent space using a few key \emph{centroids} by clustering on the hypersphere. Next, we introduce a Positive Hallucination (PosHal) module. Na\"ively exploring the latent space might lead to sub-optimal positives or even negatives. We overcome this by proposing an objective function to define hard positives. The intuition is that the similarity of the generated positives and real positives should be minimized such that both have identical closest prototypes. Since solving this optimization problem for each step of training could be expensive, we propose relaxations that let us derive a closed-form expression to find the hardest positive along a particular direction defined by a randomly selected prototype. The final solution involves a spherical linear interpolation between the anchor and a randomly selected prototype with explicit control of hardness of the generated positives. 

We experimentally verify the efficacy of \ours{} approach by experiments on standard benchmark datasets: NTU RGB-D 60, NTU RGB-D 120, and PKU-MMD II and notice consistent improvements over state-of-the-art. For example, on the linear evaluation protocol, we obtain +2.3\%, +2.3\%, and +4.5\% for cross-subject splits of the NTU-60, NTU-120, and PKU-II datasets respectively. Using our module with single-modality training leads to consistent improvements as well. Our model trained on single modality, obtains results competitive to a recent approach~\cite{mao2022cmd} which uses multiple modalities during training while being 2.5x faster. 

In summary, the following are our main contributions:
\begin{enumerate}
    \itemsep0em 
    \item We propose a new approach, \ours{} which hallucinates latent positives for use in a skeleton-based CL framework. To the best of our knowledge, we are the first to analyze the generation of positives for CL in latent-space.
    \item We define an objective function that optimizes for generating hard positives. To enable fast training, we propose relaxations to the objective which lets us derive closed-form solutions. Our approach allows for easy control of the hardness of the generated positives. 
    \item We obtain consistent improvements over the state-of-the-art methods on all benchmark datasets and tasks. Our approach is easy to use and works in uni-modal and multi-modal training, bringing benefits in both settings. 
\end{enumerate}

\section{Related Work}
\label{sec:related}

\noindent\textbf{Self-Supervised Learning:} Much of the progress in representation learning has been in the supervised paradigm. But owing to huge annotations costs and an abundance of unlabeled data, pretraining models using self-supervised techniques have received a lot of attention lately. Early works in computer vision developed pretext tasks such as predicting rotation~\cite{gidaris2018unsupervised}, solving jigsaw puzzle~\cite{noroozi2016unsupervised}, image colorization~\cite{zhang2016colorful}, and temporal order prediction~\cite{misra2016shuffle} to learn good features. Contrastive learning~\cite{gutmann2010noise,hadsell2006dimensionality} is one such pretext task that relies on instance discrimination - the goal is to classify a set of positives (augmented version of the same instance) against a set of unrelated negatives which helps the model learn good features. Recent works have demonstrated exceptional performance using these techniques in a variety of domains~\cite{jaiswal2020survey}, including images, videos, graphs, text, etc. SimCLR~\cite{chen2020simple} and MoCo~\cite{he2020momentum} frameworks have been very popular due to their ease of use and general applicability. Recently, several non-contrastive learning approaches like SwAV~\cite{caron2020unsupervised}, DINO~\cite{caron2021emerging}, MAE~\cite{he2022masked} have shown promising performance but CL still offers complementary benefits \cite{Koppula2022WhereSI, Li2022UnderstandingCI}. In this work, we focus on CL objectives which have been shown to be superior to non-contrastive ones for the task of skeleton-based SSL~\cite{mao2022cmd}. 

\noindent\textbf{Self-Supervised Skeleton/Pose-based Action Recognition:}
Supervised Skeleton-based action recognition has received a lot of attention due to its wide applicability. Research in this field has led to new models~\cite{yan2018spatial,duan2022revisiting} and learning representations~\cite{choutas2018potion}. Several works have been proposed to explore the benefits of self-supervision in this space. Some prior works have used novel pretext tasks to learn representations including skeleton colorization~\cite{yang2021skeleton}, displacement prediction~\cite{kim2022global}, skeleton inpainting~\cite{zheng2018unsupervised}, clustering~\cite{su2020predict} and Barlow twins-based learning~\cite{zhang2022skeletal}. 
Another area of interest has been on how to best represent the skeletons for self-supervised learning. Some works have explored better encoders and training schemes for skeletal sequences like Hierarchical transformers for better spatial hierarchical modeling~\cite{cheng2021hierarchical}, local-global mechanism and better handling of multiple persons in the scene~\cite{kim2022global} or use of multiple pretext tasks at different levels~\cite{chen2022hierarchically}. Complementary to innovations in the modeling of skeletons, various works have used ideas from contrastive learning for self-supervised learning of skeleton representations~\cite{rao2021augmented,thoker2021skeleton,li20213d,mao2022cmd} and have shown exceptional performance. Augmentations are crucial to contrastive learning and various works~\cite{guo2022contrastive,thoker2021skeleton} have studied techniques to augment skeletal data. Others have explored the use of additional information during training like multiple skeleton representations~\cite{thoker2021skeleton}, multiple skeleton modalities~\cite{li20213d,mao2022cmd}, local-global correspondence and attention~\cite{kim2022global}. We work with CL, owing to simplicity and strong representations. We use the same training protocols and encoders as CMD~\cite{mao2022cmd}, but show how our proposed approach can hallucinate latent positives which can be generated using a very fast plug-and-play module. Our approach shows significant improvements on both single-modality training and multi-modal training. 

\noindent\textbf{Role of positives and negatives for CL}
Prior work has shown that the use of a large number of negatives~\cite{he2020momentum} and hard negatives play an important role in contrastive learning while false negatives impact learning. DCL~\cite{chuang2020debiased} helps account and correct for false negatives, while HCL~\cite{robinson2020contrastive} also allows for control of hard negatives. MMCL~\cite{shah2022max} uses a max-margin framework to handle hard and false negatives while~\cite{ge2021robust} uses texture-based negative samples to create hard negatives. While there has been some focus on generating better positives for SSL, most approaches only consider generating better input views or to generate hard negatives. Mixup~\cite{zhang2017mixup}, Manifold mixup~\cite{verma2019manifold}, and their variants like CutMix~\cite{yun2019cutmix} have been very popular in supervised learning setups to regularize and augment data. Various approaches have proposed novel approaches to use these ideas in a self-supervised setting. M-Mix~\cite{zhang2022m} uses adaptive negative mixup in the input image space, multi-modal mixup~\cite{so2022multi} generates hard negatives by using multimodal information, ~\cite{shen2022mix} makes the model aware of the soft similarity between generated images to learn robust representations. In contrast, i-mix~\cite{lee2020mix} creates a virtual label space by training a non-parametric classifier. Closely related to our approach is MoCHI~\cite{kalantidis2020hard} which creates hard negatives for a MoCo-based framework by mixup of latent space features of positives and negatives. Unlike these works, our focus in this work is to create \emph{positives}. Instead of relying on heuristics, we instead pose generation of hard positives as an optimization problem and propose relaxation to solve this problem efficiently. 

\section{Method}
\label{sec:method}
\begin{figure}[tp]
    \centering
    \includegraphics[trim={0.5cm 0cm 0.5cm 0cm},scale=1]{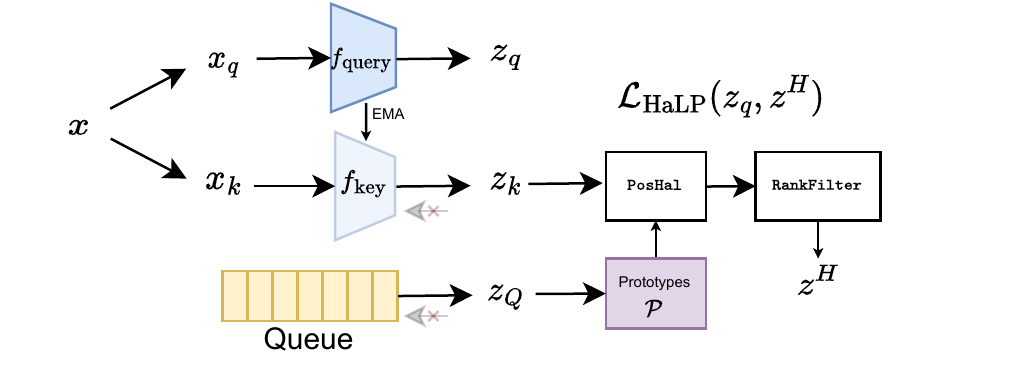}
    \caption{Overall approach to hallucinate positives. We work with a MoCo-based framework. Query and Key encoders represent the input skeleton sequence in the latent space. The queue is maintained by using past key features. The \texttt{PosHal} module hallucinates positives using the anchor and prototypes extracted from the queue. Positives satisfying the rank filter constraint are retained and used to calculate the HaLP loss. The model is trained with a weighted combination of the standard CL loss and HaLP loss.}
    \label{fig:main_model}
\end{figure}

In this work, we hallucinate new positives and use them within a self-supervised learning pipeline to learn better representations for skeleton-based action recognition.  
\cref{fig:main_model} presents an overview of our training pipeline. 

\subsection{Preliminaries}
\label{sec:preliminiaries}

\noindent\textbf{Problem setup.}
We are given a dataset $\mathcal{D}$ of unlabeled 3D skeleton sequences. We wish to learn without labels, an encoder \encoder to represent a sequence from this dataset in a high dimensional space such that the encoder can then later be adapted to a task where little training data is available. Specifically, let the skeleton sequence be \inpske~$ \in \mathbb{R}^{3 \times F \times M \times P}$, where $F$ denotes the number of frames in the sequence, $M$ is the number of joints, $P$ denotes the number of people in the scene, and the first dimension encodes the coordinates $[\text{x},\text{y},\text{z}]$. The encoder \encoder takes this sequence and generates a representation that is used for the downstream task. Prior work~\cite{mao2022cmd,thoker2021skeleton} in skeleton-based SSL has worked with various modalities like bones, joints, and motion, which are extracted from raw joint coordinates. In contrast, our approach can be applied to both single-modality and multi-modality training. 
\\

\noindent\textbf{Contrastive learning preliminaries.}
The key idea in CL is to maximize the similarity of two views (\textit{positives}) of a skeleton sequence of a video and push it away from features of other sequences (\textit{negatives}). The similarities are computed in a high-dimensional representation space. We adopt the MoCo framework~\cite{he2020momentum,chen2020improved}, which has been successfully applied to various image, video, and skeleton self-supervised learning tasks. The input skeleton sequence \inpske is first transformed by two different random cascade of augmentations $T_q, T_k \in \mathcal{T}$ to generate positives: query \xq~ and key \xk. MoCo uses separate encoders, query encoder \encoderq and key encoder \encoderk to generate L2-normalized representations \zq, \zk $\in \mathbb{R}^{D}$ on the unit hypersphere. While the \encoderq is trained using backpropagation, \encoderk~ is updated through a moving average of weights of \encoderq. In addition to encoding the positives, MoCo maintains a queue \queue $= \{z^{Q}_i\}$ with $z^{Q}_i \in \mathbb{R}^{D}$. The queue contains $L$ encoded keys from the most recent iterations of training. The elements of the queue are used as negatives in the contrastive learning process. To train the model, an InfoNCE objective function is used to update the parameters of \encoderq. The loss function used is given below: 

\begin{equation}
\label{eq:mocoloss}
	\begin{aligned}
		\mathcal{L}_{\text{CL}}=
		-\log \frac{\exp (z_q ^\top z_k / \tau)}{\exp (z_q ^\top z_k / \tau)+\sum_{i=1}^{L} \exp \left(z_q ^\top z^{Q}_i / \tau \right)},
	\end{aligned}
\end{equation}

\noindent where $\tau$ is the temperature hyperparameter. As discussed in \cref{sec:related}, there has been a lot of work in generating, and finding hard negatives which help in learning better representations. Choosing the right augmentations is critical to the learning process and works in the past have shown that including multiple views~\cite{caron2020unsupervised,caron2021emerging} helps in the learning process. But, these works craft new positives in the input space which can significantly increase the training time due to the additional forward/backward passes to the encoder. 

\subsection{Hallucinating Positives}
\label{sec:halp}
We now present our lightweight module, which can hallucinate new positives in the feature space. This has two-fold advantages: 1) This can reduce the burden on designing new data augmentations, 2) Since we hallucinate positives directly in the feature space, we do not need to backpropagate their gradients to the encoder, thus saving on expensive forward/backward passes during training. This is especially amenable to the MoCo framework~\cite{he2020momentum} where \zk does not have any gradients associated with it. Our newly generated positives \zg play the same role as \zk, except that these are hallucinated synthetic positives instead of being obtained through a real skeleton sequence.

Analogous to hard negatives~\cite{kalantidis2020hard,shah2022max,robinson2020contrastive}, we define hard positives as samples which lie far from an anchor positive in latent space but have the same semantics. Thus, we desire that our hallucinated positives should be diverse with varying amount of hardness to provide a good training signal. Further, they should have a high semantic overlap with the real positives. These are two conflicting requirements. Therefore, it is important to achieve a balance between the difficulty and similarity to the original positives when generating positive data points. This will ensure that the generated points remain valid true positives and do not introduce false positives that may hinder the training process. 

Our key intuition behind generating synthetic positives is that given the current encoded (anchor) key \zk, we can explore the high dimensional space around it to find locations that can plausibly be reached by the encoder for closely related skeleton sequences. 

We define $\text{\protoset} = \{P_1,\cdots,P_N\}$ as N cluster centroids of the data. Based on our desiderata, we can formulate the following objective to find hard positives :

\begin{equation}
\label{eq:rank_filter}
	\begin{aligned}
	&z^* = \arg\min_{z\in\mathbb{S}^{D-1}} ~\text{sim}(z,P^*_{z_k}) \\
	&\text{s.t.}\quad \text{sim}(z,P^*_{z_k}) \geq \text{sim}(z,P), \forall P\in\mathcal{P}\backslash\left\{P^*_{z_k}\right\}, 
	\end{aligned}
\end{equation}

\noindent where $\text{sim}()$ is the cosine similarity, $\mathbb{S}^{D-1}$ represents the unit hypershpere, and $P^*_{z_k} = \arg\max_{P \in \mathcal{P}} \text{sim}(z_k,P)$ is the prototype closest to \zk (our anchor). 

\begin{figure}[tp]
    \centering
    \includegraphics[trim={0cm 0cm 0cm 0cm},clip,width=\linewidth]{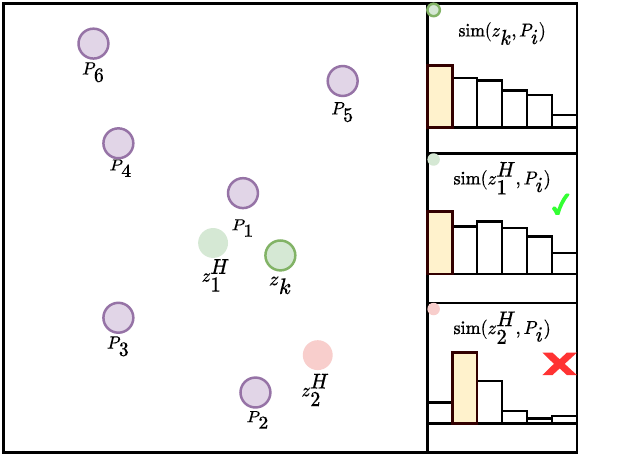}
    \caption{Intuition behind \cref{eq:rank_filter}. Our objective function enforces the constraint that the hallucinated positives and the original anchor (\zk) have the same closest prototype ($P_1$) while minimizing the similarity to \zk. For example, $z_2^H$ does not satisfy the constraint while $z_1^H$ does.}
    \label{fig:rank_filter}
\end{figure}

Intuitively, we want to generate hard positives which are far from the anchor but have the same closest prototype as the anchor. We call the constraint our Rank Filter which is visualized in \cref{fig:rank_filter}.  

Note that one could use Riemannian optimization solvers~(e.g., PyManOpt~\cite{townsend2016pymanopt}) to solve this problem. However, these are iterative and we desire quick solutions as such optimization needs to be done on every data sample. We propose simplifications to the above objective in the following subsections towards deriving a computationally cheap closed-form solution.

\subsection{Restricting the search space}
In this section, we make some simplifications to the objective in \cref{eq:rank_filter}. First, we define a new positive with the following equation: 
\begin{equation}
\label{eq:posgen}
	\begin{aligned}
        z &= \text{proj}(\text{\zk} +\text{\direction}),
	\end{aligned}
\end{equation}
 
\noindent where \zk is an anchor view to generate the positive, $\text{\direction} \in \mathbb{R}^{D}$ is the step taken, and $\text{proj}(z') = z'/\|z'\|$. The normalization step ensures that the generated point lies on the unit hypersphere (like \zk and \zq). 
To constrain the search space for the hallucinated positive, we propose to restrict \direction towards one of the prototypes \selectedproto selected at random.
This helps us relax the search space while moving towards parts of the space which are occupied by instances from the dataset which can help generate hard positives. Thus, we restrict the search space instead of searching for a $z$ (as in \cref{eq:rank_filter}). We borrow ideas from Manifold Mixup~\cite{verma2019manifold} to define intermediate points along the geodesic joining \zk and \selectedproto. 
Since we are working with points on the hypersphere, we have the following search space:
\begin{equation}
\label{eq:slerp}
	\begin{aligned}
	\text{\direction}(t,\text{\selectedproto},\text{\zk}) = \frac{\sin(1-t)\Omega}{\sin\Omega}\text{\zk} + \frac{\sin(t\Omega)}{\sin\Omega}\text{\selectedproto} - \text{\zk}, \\
	\text{where}~ t \in [0,1], \cos\Omega = \text{\selectedproto}^{\top}\text{\zk}~\text{and}~ \Omega \in [0,\pi]
	\end{aligned}
\end{equation}
We now obtain hard positive using $z^* = \text{\zk} + \text{\direction}(t^*,\text{\selectedproto},\text{\zk})$. The optimal value $t^*$ is calculated by the following modified objective function on the restricted search space. 

\begin{equation}
\label{eq:rank_filter_with_dt}
	\begin{aligned}
	&t^* = \arg\min_{t\in[0,1]} \text{sim}(z, P^*_{z_k}), \text{where} \\
	&z = \text{\zk}+d(t,\text{\selectedproto},\text{\zk}) \\
	&\text{s.t.} \quad~\text{sim}(z,P^*_{z_k}) \geq \text{sim}(z,P_j), P_j \in \mathcal{P}
	\end{aligned}
\end{equation}

While~\cref{eq:rank_filter_with_dt} restricts the search space, this still involves solving an optimization problem for each point. Next, we make another simplifying approximation. 

\subsection{Optimal solutions using pair of prototypes}
\label{optimal_sol}
Instead of solving for the  ranking objective during optimization, we solve for the following 

\begin{equation}
\label{eq:closed_form_filter}
	\begin{aligned}
	&t^* = \arg\min_{t \in [0,1]} ~\text{sim}(z,P^*_{z_k}), \text{where} \\
	&z = \text{\zk}+d(t,\text{\selectedproto},\text{\zk}) \\
    &\text{s.t.} \quad~\text{sim}(z,P^*_{z_k}) \geq \text{sim}(z,\text{\selectedproto}), \\
\end{aligned}
\end{equation}

This modified objective effectively solves \cref{eq:rank_filter_with_dt} assuming just two prototypes $\{\text{\selectedproto},\text{\closestproto}\}$. This modification lets us derive a closed-form solution to this equation. 

\begin{equation}
\label{eq:closed_form_solution}
	\begin{aligned}
	t^* &= \frac{1}{\Omega}\arctan(\frac{\sin\Omega}{\kappa+\cos\Omega}), \text{where} \\
	\kappa &= \frac{1 - \text{\selectedproto}^{\top}\text{\closestproto}}{\text{\zk}^{\top}(\text{\closestproto} - \text{\selectedproto)}},
	\end{aligned}
\end{equation}

Intuitively, $t^*$ restricts the part of the geodesic between \zk and \selectedproto which are good positives. 
We then generate the new positives as $\text{\zg} = \text{\zk} + d(t_{c},\text{\selectedproto},\text{\zk})$ where $t_c \sim \texttt{uniform}(0, \lambda t^*)$ where the fixed scalar $\lambda$ allows us to control the level of hardness of the generated positives. \texttt{PosHal}~(\cref{fig:main_model}) uses this strategy to efficiently generate positives for a given batch. Since our relaxation \cref{eq:closed_form_filter} considers only two prototypes, the generated positives might not satisfy the original ranking constraint~(\cref{eq:rank_filter}). Thus, we pass the generated positives through the rank-filter to obtain the final set of filtered positives (\cref{fig:main_model}). 

\subsection{How to obtain prototypes }

The prototypes $\mathcal{P}$ could intuitively represent various classes, action attributes, etc. Since we are working with points on the unit hypersphere, $\mathbb{S}^{D-1}$, we propose to use k-Means clustering on the hypersphere manifold with the associated Riemannian metric. Since the queue gets updated in a first-in-first-out fashion, we use the \texttt{topK} most recent elements of the queue to obtain the prototypes. Next, we calculate the similarities of the \zk to each of the prototypes.

\begin{algorithm}[!t]
\caption{\label{alg:main_algo} HaLP : Halucinating Latent Positives}
\begin{algorithmic}
    \STATE \textbf{Input:} \xk, \xq, \encoderk, \encoderq, queue \queue
    \STATE \textbf{Output:} $\mathcal{L}_{\text{HaLP}}$ 
    \STATE \textcolor{gray}{\# Extract key and query features}
    \STATE $z_k, z_q = E_k(x_k), E_q(x_q)$
    \STATE \textcolor{gray}{\# Cluster most recent queue elements into N prototypes}
    \STATE $\mathcal{P}=\texttt{sphere\_cluster}(\texttt{topK}(Q))$ 
    \STATE \textcolor{gray}{\# Find the closest prototype to key}
    \STATE $P^*_{z_k} = \arg\max_{P} \{\text{sim}(z_k,P_j)\}_j$
    \STATE \textcolor{gray}{\# Select a prototype to step towards}
    \STATE ${P}_{\text{sel}}=\texttt{random}(\mathcal{P})$
    \STATE \textcolor{gray}{\# Determine how far we can move from key while still being a member of $P^*_{z_k}$}
    \STATE $t^*$ using \cref{eq:closed_form_solution}
    \STATE \textcolor{gray}{\# Control hardness using $\lambda$ }
    \STATE $t_c \sim \texttt{uniform}(0, \lambda t^*)$
    \STATE \textcolor{gray}{\# Generate hallucinated positive of $z_k$ (\cref{eq:slerp})} 
    \STATE $z_i^{'H} = z_k + d(t_c,\text{\selectedproto},\text{\zk})$ 
    \STATE \textcolor{gray}{\# Apply rank-filter to discard generated positives not satisfying constraint in \cref{eq:rank_filter}}
    \STATE $z_{i}^{H} = \texttt{rank\_{filter}} (z_{i}^{'H})$
    \STATE \textcolor{gray}{\# Compute $\mathcal{L}_{\text{HaLP}}$ using \cref{eq:finalloss}}
    
    \STATE return $\mathcal{L}_{\text{HaLP}}$
\end{algorithmic}
\end{algorithm}

\subsection{Loss function}
The final step in the proposed approach involves using these generated points to train the model. Note that these generated points do not have any gradient through them since in \cref{eq:posgen}, both \zk and \direction are detached from the computational graph. Thus, we use the generated points as \emph{hallucinated keys} in the MoCo framework. 
We train our models using the following weighted loss
\begin{equation}
\label{eq:finalloss}
	\begin{aligned}
		\mathcal{L}_{\text{total}} &= \mathcal{L}_{\text{CL}} + \mu\mathcal{L}_{\text{HaLP}}~~\text{where}\\
		\mathcal{L}_{\text{HaLP}} &= -\frac{1}{G_{\text{filtered}}} \sum_i^{G_{\text{filtered}}} z_q ^\top \text{\zg} / \tau
	\end{aligned}
\end{equation}
where $G_{\text{filtered}}$ is the number of filtered positives. \cref{alg:main_algo} summarizes our entire approach. 

\section{Experiments}
\label{sec:experiments}

\begin{table}[t!]
	\centering
		\caption{Results on linear evaluation: Our proposed module HaLP achieves state-of-the-art results and improves the performance on NTU-60 x-sub dataset by 2.3\%, NTU-120 x-sub by 2.3\% and PKU-II by 4.5\%. We also show significant improvements over the single modality Baseline. Our proposed model is a lightweight module, with minimal overheads and helps in achieving strong performances across various datasets.}
		\resizebox{1.0\linewidth}{!}{
			\begin{tabular}{lccccc}
				\toprule
				\multirow{2}{*}{Method}  & \multicolumn{2}{c}{\textbf{NTU-60}}	& \multicolumn{2}{c}{\textbf{NTU-120}}	& \textbf{PKU-II} \\
				\cmidrule(lr){2-3} \cmidrule(lr){4-5} \cmidrule(lr){6-6}
				&  						 x-sub          & x-view  			& x-sub         	& x-set       & x-sub\\
				\midrule
				\multicolumn{6}{l}{\textit{Additional training modalities or encoders}} \\
				ISC \cite{thoker2021skeleton}		     				& 76.3           & 85.2   			& 67.1          & 67.9 				& 36.0\\
				CrosSCLR-B ~\cite{mao2022cmd} 	                     				& 77.3           & 85.1   			& 67.1          & 68.6 				& 41.9\\
				CMD ~\cite{mao2022cmd}             				& 79.8  & 86.9	& 70.3 & 71.5 	& 43.0\\
				\textbf{\ours{} + CMD}               				& \textbf{82.1} & \textbf{88.6}	& \textbf{72.6} & \textbf{73.1}	& \textbf{47.5}\\
				\hline
				\multicolumn{6}{l}{\textit{Training using only joint}} \\
				LongT GAN \cite{zheng2018unsupervised}   				& 39.1           & 48.1   			& -             & - 				& 26.0\\
				MS$^2$L \cite{Lin2020MS2LMS}               				& 52.6           & -      			& -             & - 				& 27.6\\
				P\&C \cite{su2020predict}		         				& 50.7           & 76.3       		& 42.7          & 41.7 				& 25.5\\
				AS-CAL \cite{rao2021augmented} 			 				& 58.5           & 64.8   			& 48.6          & 49.2 				& -\\
				H-Transformer \cite{cheng2021hierarchical} & 69.3 & 72.8 & - & - & -\\
				SKT~\cite{zhang2022skeletal} & 72.6 & 77.1 & 62.6 & 64.3 & - \\
				GL-Transformer~\cite{kim2022global} & 76.3 & 83.8 & 66.0 & 68.7 & - \\
				SeBiReNet \cite{Nie2020Unsupervised3H}     				& -              & 79.7				& -      		& - 				& -\\
				AimCLR \cite{guo2022contrastive}	             				& 74.3           & 79.7   			& -             & - 				& -\\
				Baseline        				& 78.0 & 85.5	& 69.1 & 69.8	& 42.9\\
				\textbf{\ours{}}               				& \textbf{79.7} & \textbf{86.8}	& \textbf{71.1} & \textbf{72.2} 	& \textbf{43.5}\\
				\bottomrule
			\end{tabular}
		}
\label{tab:lineval_comparison}
\end{table}
\subsection{Datasets:}
\noindent\textbf{NTU RGB+D 60:} NTU-60 is a large-scale action recognition dataset consisting of 56880 action sequences belonging to 60 categories performed by 40 subjects. The dataset consists of multiple captured modalities, including appearance, depth and Kinect-v2 captured skeletons. Following prior work in skeleton-based action recognition, we only work with the skeleton sequences. These sequences consist of 3D coordinates of 25 skeleton joints. The dataset is used with two protocols: Cross-subject: where half of the subjects are present in the training and the rest for the test; Cross-View where two of the camera's views are used for training and the rest for the test. 
\\

\noindent\textbf{NTU RGB+D 120:} This dataset was collected as an extension of the NTU RGB+D 60 dataset. This dataset extends the number of actions from 60 to 120, and the number of subjects from 40 to 10 and includes a total of 32 camera setups with varying distances and backgrounds. The dataset contains 114480 sequences. In addition to a Cross-Subject protocol, The dataset proposes a cross-setup protocol as a replacement for cross-view to account for changes in camera distance, views, and backgrounds. 
\\

\noindent\textbf{PKU MMD - II:} is a  benchmark multi-modal 3D human action understanding dataset. Following prior works, we work with the phase 2 dataset which is made quite challenging due to large view variations. Like NTU-60/120, we work with the Kinect-v2 captured sequences provided with the dataset. We use the cross-subject protocol, which has 5332 sequences for the train and 1613 for the test set.  
\\

\subsection{Implementation details}
\noindent\textbf{Encoder models \encoder:} For a fair comparison with recent state-of-the-art~\cite{mao2022cmd}, we make use of a Bidirectional GRU-based encoder. Precisely, it consists of 3D bidirectional recurrent layers with a hidden dimension of 1024. The encoder is followed by a projection layer, and L2 normalized features from the projection layer are used for self-supervised pre-training. As is common in self-supervised learning, pre-MLP features are used for the downstream task. Authors in~\cite{mao2022cmd} have shown the benefit of a BiGRU encoder over graph convolutional networks or transformer-based models. We use the same preprocessing steps as prior work where a maximum of two actors are encoded in the sequence, and inputs corresponding to the non-existing actor are set to zero. Inputs to the model are temporally resized to 64 frames. 
\\

\noindent\textbf{Baseline:} For a fair comparison with prior works using a single modality during training, we also compare our approach to `baseline'. This essentially implements a single modality, vanilla MoCo using CMD~\cite{mao2022cmd} framework without the cross-modal distillation loss~\cite{mao2022cmd} or the hallucinated positives. This baseline demonstrates the benefit of CL for this problem using the joint modality alone.
\\

\noindent\textbf{Multimodal training}: Our plug-and-play approach is not dependent on the modalities used during training. Unless otherwise specified, we use a single modality (Joint) during training. For multi-modality training experiments, we use the CMD~\cite{mao2022cmd} framework and hallucinate per-modality positives. We call this approach~\ours{} + CMD. Note that using cross-modal positives could lead to further improvements, but we leave this extension to future work to keep our approach general. 
\\

We follow the same pre-training hyperparameters as ISC~\cite{thoker2021skeleton,mao2022cmd}. The models are trained with SGD with a momentum of 0.9 and weight decay of 0.0001. We use a batch size of 64 and a learning rate of 0.01. Models are pretrained for 450 and 1000 epochs on NTU-60/120 and PKU, respectively. The queue size used is 16384. We use the same skeleton augmentations as ISC and CMD. These include pose jittering, shear augmentation and temporal random resize crop. 
\\

\noindent\textbf{\ours{} specific implementation details:}
We generate 100 positives per anchor. We use \texttt{Geomstats}~\cite{geomstats} for k-Means clustering on the hypersphere with a tolerance of 1E-3 and initial step size of 1.0. The prototypes are updated every five iterations. 256 ($K$) most recent values of the queue were used to cluster. $\lambda=0.8$ is used for all experiments. We use $\mu=0$ for the first 200 epochs and $\mu=1$ for the rest. Wandb~\cite{wandb} was used for experiment tracking. 
\\

\noindent\textbf{Evaluation protocols:}
We evaluate the models on three standard downstream tasks: Linear evaluation, kNN evaluation and transfer learning. In \cref{sec:sota_comparison} we describe the protocols followed by our results. Additional implementation details and experiments are present in the supplementary material. 

\subsection{Comparison with state-of-the-art}
\label{sec:sota_comparison}
\noindent\textbf{Linear Evaluation:} Here, we freeze the self-supervised pre-trained encoder and attach an MLP classifier head to it. The model is then trained with labels on the dataset. We pretrain our model on the NTU-60, NTU-120 and PKU-II datasets. We present our results in Table~\ref{tab:lineval_comparison}. We see that in both uni-modal and multi-modal training our approach outperforms the state-of-the-art on this protocol which shows that our approach learns better features. Further, it is interesting to see that apart from outperforming all single modality baselines, training using \ours{} on a single modality even shows competitive performance to models trained using multiple modalities. 
\\

\noindent\textbf{kNN Evaluation:} This is a hyperparameter-free \& training-free evaluation protocol which directly uses the pre-trained encoder and applies a k-Nearest Neighbor classifier (k=1) to the learned features of the training samples. 
We present our results using this protocol in Table~\ref{tab:knn_comparison}. We again see considerable improvements on the various task settings showing that our approach works better even in a hyperparameter-free evaluation setting.  
\\

\begin{table}[t!]
	\centering
		\caption{kNN evaluation: In addition to linear probing, we also show improved performances on kNN evaluation. Similar to linear probing our proposed module HaLP leads to significant performance improvements on both NTU-60 and NTU-120 datasets compared to the single-modality baseline. Using our approach with CMD~\cite{mao2022cmd} leads to further gains over state-of-the-art.}
\begin{tabular}{lcccc}
			\toprule
			\multirow{2}{*}{Method}   		  & \multicolumn{2}{c}{\textbf{NTU-60}}	& \multicolumn{2}{c}{\textbf{NTU-120}}\\
			\cmidrule(lr){2-3} \cmidrule(lr){4-5}
			& x-sub          & x-view  			& x-sub         	& x-set\\
			\midrule 
			\multicolumn{5}{l}{\textit{Additional training modalities or encoders}} \\
			ISC \cite{thoker2021skeleton}		    & 62.5           & 82.6   			& 50.6          & 52.3 \\
			CrosSCLR-B  		    & 66.1           & 81.3   			& 52.5          & 54.9 \\
			CMD             & 70.6  & 85.4	& 58.3 & 60.9 \\
			\textbf{\ours{}+CMD}             &  \textbf{71.0} & \textbf{86.4}	& \textbf{59.4} & \textbf{61.9} \\
			\hline
			\multicolumn{5}{l}{\textit{Additional training modalities or encoders}} \\
			LongT GAN \cite{zheng2018unsupervised}  & 39.1           & 48.1   			& 31.5             & 35.5 \\
			P\&C \cite{su2020predict}              & 50.7           & 76.3      			& 39.5             & 41.8 \\ 			
			Baseline    & 63.6  & 	82.8 &  51.7 & 55.3 \\
			\textbf{\ours{}}             & \textbf{65.8}  & \textbf{83.6}	& \textbf{55.8} & \textbf{59.0} \\
			\bottomrule
		\end{tabular}
		\label{tab:knn_comparison}
\end{table}

\noindent\textbf{NTU to PKU transfer:} Here, we evaluate whether the pre-trained models trained on NTU can be used to transfer to PKU, which has much less training data. A classifier MLP is attached to an NTU pre-trained encoder, and the entire model is finetuned with labels on the PKU dataset. In Table~\ref{tab:pku_transfer}, we see that our approach improves over the state-of-the-art, showing more transferable features. 
\begin{table}[t!]
	\centering
		\caption{NTU to PKU transfer: We adapt NTU pretrained models to the PKU-II dataset. We observe that our model improves in the transfer learning setup as well.}
\begin{tabular}{lcccc}
			\toprule
			\multirow{2}{*}{Method}   		  & \multicolumn{2}{c}{\textbf{To PKU-II}} \\
			\cmidrule(lr){2-3}
			 & NTU-60          & NTU-120 \\
			\midrule
			\multicolumn{3}{l}{\textit{Additional training modalities or encoders}} \\
			ISC \cite{thoker2021skeleton}    		 & 51.1           & 52.3\\
			CrosSCLR-B		    & 54.0           & 52.8 \\
			CMD               & 56.0  & 57.0\\
			\textbf{\ours{} + CMD} & \textbf{56.6} & \textbf{57.3}  \\
			\hline
			\multicolumn{3}{l}{\textit{Training using only joint}} \\
			LongT GAN \cite{zheng2018unsupervised}	 & 44.8           & -\\
			MS$^2$L \cite{Lin2020MS2LMS}     		     & 45.8           & -\\
Baseline               & 53.3& 53.4\\
			\textbf{\ours{}}               & \textbf{54.8} & \textbf{55.4}\\
			\bottomrule
		\end{tabular}
		\label{tab:pku_transfer}
\end{table}

\subsection{Analyses and ablations}
Next, we present ablation analyses on our models. In this section, we work with the single-modality training of \ours{} for NTU-60 Cross Subject split unless otherwise mentioned. Models for these experiments are pre-trained on NTU-60 using the cross-subject protocol, and results for linear evaluation are presented. 
\\

\noindent\textbf{Training time and Memory requirements:} While our approach involves clustering on the hypersphere manifold, we empirically show that our approach incurs only marginal overheads compared to the baseline (\cref{tab:timing_and_memory}). We see that \ours{}, has performance comparable to CMD while incurring much less training memory and training time penalty, which makes use of multiple modalities during training. Further, \ours{}+CMD is comparable to CMD in training time and memory requirements but outperforms CMD~\cite{mao2022cmd}. Note that since our technique modifies the loss, the inference pipeline remains exactly the same as the baseline, and we run at 1x the time of the baseline. 
\begin{table}[t]
\centering

\caption{Computational overhead: Our proposed module HaLP is lightweight and results in very small computational overheads and can potentially be added to any contrastive learning method.}
\resizebox{\columnwidth}{!}{
\begin{tabular}{lccc}
\toprule
Method & Time/epoch & Train GPU memory & NTU-60 x-sub \\
\midrule
Baseline & 1x & 1x & 78.0 \\
\ours{} & 1.13x & 1x & \textbf{79.7} \\
\hline
CMD & 3x & 1.94x & 79.8 \\
\ours{}+CMD & 3.32x & 1.94x & \textbf{82.1} \\
\bottomrule
\end{tabular}}

\vspace{-0.05in}
\label{tab:timing_and_memory}
\end{table}

\\

\noindent\textbf{Can query-key be used to hallucinate positives? } One potentially simple way to generate positives would be to interpolate between \zk and \zq along the geodesic joining them. We observe that this approach does not improve over the baseline and has a performance of 77.92\%. This shows that naively selecting directions to explore may not lead to better training. 
\\

\noindent\textbf{Ablation on $\lambda$:} 
Recollect that in \cref{optimal_sol}, we choose $t_c \sim \texttt{uniform}(0, \lambda t^*)$. The value of $\lambda$ effectively controls the maximum hardness of the generated positives. 
In \cref{tab:lambda_ablation}, we modify $\lambda$, and observe that a value of 1.0 and 0.8 works well. We found that the $\lambda = 0.8$ works consistently well across all datasets and use it for all our experiments. Instead of generating positives with a range of hardness, one could also just use the positive defined by $t_c = t^*$. We find this is suboptimal (79.4\%). Thus, the use of only the hardest positives is not very effective.
\\

        \begin{table}[t]
        \centering
\caption{
Effect of changing $\lambda$, which controls the maximum hardness of generated points. We find that that using a large value of $\lambda$ (harder positives) shows better performance.}
\resizebox{\linewidth}{!}{%
        \begin{tabular}{lcccccc}
        \toprule
        $\lambda \rightarrow$ & 1 & 0.8 & 0.6 & 0.4 & 0.2 \\
        \midrule
        NTU-60 x-sub & \textbf{79.7} & \textbf{79.7} & 79.6 & 79.6 & 79.5 \\
        \bottomrule
        \end{tabular}
}

\label{tab:lambda_ablation}
\end{table}

\noindent\textbf{How many prototypes to use?} In our approach, prototypes $\mathcal{P}$ are obtained using a k-Means clustering algorithm on the hypersphere manifold. Hallucinated latent positives are obtained by stepping along a randomly chosen prototype. In this experiment, we vary the number of clusters and their effect on the performance. In \cref{tab:numproto}, we see that using 20 prototypes is optimal for NTU-60 dataset.  
\\

        \begin{table}[t]
        \centering
\caption{
Effect of varying the number of prototypes (N) to be extracted. We observe that using 20 prototypes for NTU-60 leads to optimal performance.}
\resizebox{\linewidth}{!}{%
        \begin{tabular}{lcccc}
        \toprule
        \# Prototypes (N) $\rightarrow$ & 10 & 20 & 40 & 60  \\
        \midrule
        NTU-60 x-sub & 79.5 & \textbf{79.7} & 79.5 & 79.5 \\
        \bottomrule
        \end{tabular}
}

\label{tab:numproto}
\end{table}

\noindent\textbf{Which anchor to use? :} In our work, we choose \zk as the anchor used to hallucinate positives. An alternative could be to use \zq instead. We found that this does not improve over the baseline and has linear evaluation performance of 77.9\%. This might be due to the nature of the loss since it compares the similarity of the generated positive to \zq which might lead to trivial training signals. Based on this observation, we use \zk as an anchor for our experiments.
\\

\noindent\textbf{Supplementary material:} Additional experiments and analyses including semi-supervised learning, multi-modal ensembles, the effect of top-K, \ours{} applied to other frameworks and tasks, additional results using multi-modal training, and alternate variants are provided in the supplementary material. 
\section{Conclusion}
\label{sec:conclusion}

We present an approach to hallucinate positives in the latent space for self-supervised representation learning of skeleton sequences. We define an objective function to generate hard latent positives. On-the-fly generation of positives requires that the process is fast and introduces minimal overheads. To that end, we propose relaxations to the original objective and derive closed-form solutions for the hard positive. The generated positives with varying amounts of hardness are then used within a contrastive learning framework. Our approach offers a fast alternative to hand-crafting new augmentations. We show that our approach leads to state-of-the-art performance on various standard benchmarks in self-supervised skeleton representation learning. 

\section{Acknowledgements}
AS acknowledges support through a fellowship from JHU + Amazon Initiative for Interactive AI (AI2AI). AR and RC acknowledge support from an ONR MURI grant N00014-20-1-2787. SM and DJ were supported in part by the National Science Foundation under grant no. IIS-1910132 and IIS-2213335. AS thanks Yunyao Mao for help with questions regarding CMD. 

{\small
\bibliographystyle{ieee_fullname}
\bibliography{11_references}

\begin{thebibliography}{10}\itemsep=-1pt

\bibitem{geomstats}
Geomstats https://geomstats.github.io/.

\bibitem{wandb}
Lukas Biewald.
\newblock Experiment tracking with weights and biases, 2020.
\newblock Software available from wandb.com.

\bibitem{caron2020unsupervised}
Mathilde Caron, Ishan Misra, Julien Mairal, Priya Goyal, Piotr Bojanowski, and
  Armand Joulin.
\newblock Unsupervised learning of visual features by contrasting cluster
  assignments.
\newblock {\em Advances in Neural Information Processing Systems},
  33:9912--9924, 2020.

\bibitem{caron2021emerging}
Mathilde Caron, Hugo Touvron, Ishan Misra, Herv{\'e} J{\'e}gou, Julien Mairal,
  Piotr Bojanowski, and Armand Joulin.
\newblock Emerging properties in self-supervised vision transformers.
\newblock In {\em Proceedings of the IEEE/CVF International Conference on
  Computer Vision}, pages 9650--9660, 2021.

\bibitem{chang2011kinect}
Yao-Jen Chang, Shu-Fang Chen, and Jun-Da Huang.
\newblock A kinect-based system for physical rehabilitation: A pilot study for
  young adults with motor disabilities.
\newblock {\em Research in developmental disabilities}, 32(6):2566--2570, 2011.

\bibitem{chen2020simple}
Ting Chen, Simon Kornblith, Mohammad Norouzi, and Geoffrey Hinton.
\newblock A simple framework for contrastive learning of visual
  representations.
\newblock In {\em International conference on machine learning}, pages
  1597--1607. PMLR, 2020.

\bibitem{chen2020improved}
Xinlei Chen, Haoqi Fan, Ross Girshick, and Kaiming He.
\newblock Improved baselines with momentum contrastive learning.
\newblock {\em arXiv preprint arXiv:2003.04297}, 2020.

\bibitem{chen2022hierarchically}
Yuxiao Chen, Long Zhao, Jianbo Yuan, Yu Tian, Zhaoyang Xia, Shijie Geng, Ligong
  Han, and Dimitris~N Metaxas.
\newblock Hierarchically self-supervised transformer for human skeleton
  representation learning.
\newblock In {\em European Conference on Computer Vision}, pages 185--202.
  Springer, 2022.

\bibitem{cheng2021hierarchical}
Yi-Bin Cheng, Xipeng Chen, Junhong Chen, Pengxu Wei, Dongyu Zhang, and Liang
  Lin.
\newblock Hierarchical transformer: Unsupervised representation learning for
  skeleton-based human action recognition.
\newblock In {\em 2021 IEEE International Conference on Multimedia and Expo
  (ICME)}, pages 1--6. IEEE, 2021.

\bibitem{choutas2018potion}
Vasileios Choutas, Philippe Weinzaepfel, J{\'e}r{\^o}me Revaud, and Cordelia
  Schmid.
\newblock Potion: Pose motion representation for action recognition.
\newblock In {\em Proceedings of the IEEE conference on computer vision and
  pattern recognition}, pages 7024--7033, 2018.

\bibitem{chuang2020debiased}
Ching-Yao Chuang, Joshua Robinson, Yen-Chen Lin, Antonio Torralba, and Stefanie
  Jegelka.
\newblock Debiased contrastive learning.
\newblock {\em Advances in neural information processing systems},
  33:8765--8775, 2020.

\bibitem{corona2021meva}
Kellie Corona, Katie Osterdahl, Roderic Collins, and Anthony Hoogs.
\newblock Meva: A large-scale multiview, multimodal video dataset for activity
  detection.
\newblock In {\em Proceedings of the IEEE/CVF Winter Conference on Applications
  of Computer Vision}, pages 1060--1068, 2021.

\bibitem{duan2022revisiting}
Haodong Duan, Yue Zhao, Kai Chen, Dahua Lin, and Bo Dai.
\newblock Revisiting skeleton-based action recognition.
\newblock In {\em Proceedings of the IEEE/CVF Conference on Computer Vision and
  Pattern Recognition}, pages 2969--2978, 2022.

\bibitem{feichtenhofer2019slowfast}
Christoph Feichtenhofer, Haoqi Fan, Jitendra Malik, and Kaiming He.
\newblock Slowfast networks for video recognition.
\newblock In {\em Proceedings of the IEEE/CVF international conference on
  computer vision}, pages 6202--6211, 2019.

\bibitem{ge2021robust}
Songwei Ge, Shlok Mishra, Chun-Liang Li, Haohan Wang, and David Jacobs.
\newblock Robust contrastive learning using negative samples with diminished
  semantics.
\newblock {\em Advances in Neural Information Processing Systems},
  34:27356--27368, 2021.

\bibitem{Ghosh2021ContrastiveLI}
Aritra Ghosh and Andrew~S. Lan.
\newblock Contrastive learning improves model robustness under label noise.
\newblock {\em 2021 IEEE/CVF Conference on Computer Vision and Pattern
  Recognition Workshops (CVPRW)}, pages 2697--2702, 2021.

\bibitem{gidaris2018unsupervised}
Spyros Gidaris, Praveer Singh, and Nikos Komodakis.
\newblock Unsupervised representation learning by predicting image rotations.
\newblock {\em arXiv preprint arXiv:1803.07728}, 2018.

\bibitem{guo2022contrastive}
Tianyu Guo, Hong Liu, Zhan Chen, Mengyuan Liu, Tao Wang, and Runwei Ding.
\newblock Contrastive learning from extremely augmented skeleton sequences for
  self-supervised action recognition.
\newblock In {\em Proceedings of the AAAI Conference on Artificial
  Intelligence}, volume~36, pages 762--770, 2022.

\bibitem{Guo2022ContrastiveLF}
Tianyu Guo, Hong Liu, Zhan Chen, Mengyuan Liu, Tao Wang, and Runwei Ding.
\newblock Contrastive learning from extremely augmented skeleton sequences for
  self-supervised action recognition.
\newblock In {\em AAAI}, 2022.

\bibitem{gutmann2010noise}
Michael Gutmann and Aapo Hyv{\"a}rinen.
\newblock Noise-contrastive estimation: A new estimation principle for
  unnormalized statistical models.
\newblock In {\em Proceedings of the thirteenth international conference on
  artificial intelligence and statistics}, pages 297--304. JMLR Workshop and
  Conference Proceedings, 2010.

\bibitem{hadsell2006dimensionality}
Raia Hadsell, Sumit Chopra, and Yann LeCun.
\newblock Dimensionality reduction by learning an invariant mapping.
\newblock In {\em 2006 IEEE Computer Society Conference on Computer Vision and
  Pattern Recognition (CVPR'06)}, volume~2, pages 1735--1742. IEEE, 2006.

\bibitem{he2022masked}
Kaiming He, Xinlei Chen, Saining Xie, Yanghao Li, Piotr Doll{\'a}r, and Ross
  Girshick.
\newblock Masked autoencoders are scalable vision learners.
\newblock In {\em Proceedings of the IEEE/CVF Conference on Computer Vision and
  Pattern Recognition}, pages 16000--16009, 2022.

\bibitem{he2020momentum}
Kaiming He, Haoqi Fan, Yuxin Wu, Saining Xie, and Ross Girshick.
\newblock Momentum contrast for unsupervised visual representation learning.
\newblock In {\em Proceedings of the IEEE/CVF conference on computer vision and
  pattern recognition}, pages 9729--9738, 2020.

\bibitem{jaiswal2020survey}
Ashish Jaiswal, Ashwin~Ramesh Babu, Mohammad~Zaki Zadeh, Debapriya Banerjee,
  and Fillia Makedon.
\newblock A survey on contrastive self-supervised learning.
\newblock {\em Technologies}, 9(1):2, 2020.

\bibitem{johansson1973visual}
Gunnar Johansson.
\newblock Visual perception of biological motion and a model for its analysis.
\newblock {\em Perception \& psychophysics}, 14(2):201--211, 1973.

\bibitem{kalantidis2020hard}
Yannis Kalantidis, Mert~Bulent Sariyildiz, Noe Pion, Philippe Weinzaepfel, and
  Diane Larlus.
\newblock Hard negative mixing for contrastive learning.
\newblock {\em Advances in Neural Information Processing Systems},
  33:21798--21809, 2020.

\bibitem{kazakos2019epic}
Evangelos Kazakos, Arsha Nagrani, Andrew Zisserman, and Dima Damen.
\newblock Epic-fusion: Audio-visual temporal binding for egocentric action
  recognition.
\newblock In {\em Proceedings of the IEEE/CVF International Conference on
  Computer Vision}, pages 5492--5501, 2019.

\bibitem{kidzinski2020deep}
{\L}ukasz Kidzi{\'n}ski, Bryan Yang, Jennifer~L Hicks, Apoorva Rajagopal,
  Scott~L Delp, and Michael~H Schwartz.
\newblock Deep neural networks enable quantitative movement analysis using
  single-camera videos.
\newblock {\em Nature communications}, 11(1):1--10, 2020.

\bibitem{kim2022global}
Boeun Kim, Hyung~Jin Chang, Jungho Kim, and Jin~Young Choi.
\newblock Global-local motion transformer for unsupervised skeleton-based
  action learning.
\newblock In {\em European Conference on Computer Vision}, pages 209--225.
  Springer, 2022.

\bibitem{kong2022human}
Yu Kong and Yun Fu.
\newblock Human action recognition and prediction: A survey.
\newblock {\em International Journal of Computer Vision}, 130(5):1366--1401,
  2022.

\bibitem{Koppula2022WhereSI}
Skanda Koppula, Yazhe Li, Evan Shelhamer, Andrew Jaegle, Nikhil Parthasarathy,
  Relja Arandjelovi{\'c}, Jo{\~a}o Carreira, and Olivier~J. H'enaff.
\newblock Where should i spend my flops? efficiency evaluations of visual
  pre-training methods.
\newblock {\em ArXiv}, abs/2209.15589, 2022.

\bibitem{lee2020mix}
Kibok Lee, Yian Zhu, Kihyuk Sohn, Chun-Liang Li, Jinwoo Shin, and Honglak Lee.
\newblock i-mix: A domain-agnostic strategy for contrastive representation
  learning.
\newblock {\em arXiv preprint arXiv:2010.08887}, 2020.

\bibitem{Li2022UnderstandingCI}
Alexander Li, Alexei~A. Efros, and Deepak Pathak.
\newblock Understanding collapse in non-contrastive siamese representation
  learning.
\newblock In {\em ECCV}, 2022.

\bibitem{li20213d}
Linguo Li, Minsi Wang, Bingbing Ni, Hang Wang, Jiancheng Yang, and Wenjun
  Zhang.
\newblock 3d human action representation learning via cross-view consistency
  pursuit.
\newblock In {\em Proceedings of the IEEE/CVF Conference on Computer Vision and
  Pattern Recognition}, pages 4741--4750, 2021.

\bibitem{li2018resound}
Yingwei Li, Yi Li, and Nuno Vasconcelos.
\newblock Resound: Towards action recognition without representation bias.
\newblock In {\em Proceedings of the European Conference on Computer Vision
  (ECCV)}, pages 513--528, 2018.

\bibitem{li2019repair}
Yi Li and Nuno Vasconcelos.
\newblock Repair: Removing representation bias by dataset resampling.
\newblock In {\em Proceedings of the IEEE/CVF conference on computer vision and
  pattern recognition}, pages 9572--9581, 2019.

\bibitem{Lin2020MS2LMS}
Lilang Lin, Sijie Song, Wenhan Yang, and Jiaying Liu.
\newblock Ms2l: Multi-task self-supervised learning for skeleton based action
  recognition.
\newblock {\em Proceedings of the 28th ACM International Conference on
  Multimedia}, 2020.

\bibitem{mao2022cmd}
Yunyao Mao, Wengang Zhou, Zhenbo Lu, Jiajun Deng, and Houqiang Li.
\newblock Cmd: Self-supervised 3d action representation learning with
  cross-modal mutual distillation.
\newblock {\em arXiv preprint arXiv:2208.12448}, 2022.

\bibitem{meng2021adafuse}
Yue Meng, Rameswar Panda, Chung-Ching Lin, Prasanna Sattigeri, Leonid
  Karlinsky, Kate Saenko, Aude Oliva, and Rogerio Feris.
\newblock Adafuse: Adaptive temporal fusion network for efficient action
  recognition.
\newblock {\em arXiv preprint arXiv:2102.05775}, 2021.

\bibitem{misra2016shuffle}
Ishan Misra, C~Lawrence Zitnick, and Martial Hebert.
\newblock Shuffle and learn: unsupervised learning using temporal order
  verification.
\newblock In {\em European Conference on Computer Vision}, pages 527--544.
  Springer, 2016.

\bibitem{Nie2020Unsupervised3H}
Qiang Nie, Ziwei Liu, and Yunhui Liu.
\newblock Unsupervised 3d human pose representation with viewpoint and pose
  disentanglement.
\newblock In {\em ECCV}, 2020.

\bibitem{noroozi2016unsupervised}
Mehdi Noroozi and Paolo Favaro.
\newblock Unsupervised learning of visual representations by solving jigsaw
  puzzles.
\newblock In {\em European conference on computer vision}, pages 69--84.
  Springer, 2016.

\bibitem{rao2021augmented}
Haocong Rao, Shihao Xu, Xiping Hu, Jun Cheng, and Bin Hu.
\newblock Augmented skeleton based contrastive action learning with momentum
  lstm for unsupervised action recognition.
\newblock {\em Information Sciences}, 569:90--109, 2021.

\bibitem{robinson2020contrastive}
Joshua Robinson, Ching-Yao Chuang, Suvrit Sra, and Stefanie Jegelka.
\newblock Contrastive learning with hard negative samples.
\newblock {\em arXiv preprint arXiv:2010.04592}, 2020.

\bibitem{shah2022max}
Anshul Shah, Suvrit Sra, Rama Chellappa, and Anoop Cherian.
\newblock Max-margin contrastive learning.
\newblock In {\em Proceedings of the AAAI Conference on Artificial
  Intelligence}, volume~36, pages 8220--8230, 2022.

\bibitem{shen2022mix}
Zhiqiang Shen, Zechun Liu, Zhuang Liu, Marios Savvides, Trevor Darrell, and
  Eric Xing.
\newblock Un-mix: Rethinking image mixtures for unsupervised visual
  representation learning.
\newblock In {\em Proceedings of the AAAI Conference on Artificial
  Intelligence}, volume~36, pages 2216--2224, 2022.

\bibitem{Si2020AdversarialSL}
Chenyang Si, Xuecheng Nie, Wei Wang, Liang Wang, Tieniu Tan, and Jiashi Feng.
\newblock Adversarial self-supervised learning for semi-supervised 3d action
  recognition.
\newblock {\em ArXiv}, abs/2007.05934, 2020.

\bibitem{so2022multi}
Junhyuk So, Changdae Oh, Minchul Shin, and Kyungwoo Song.
\newblock Multi-modal mixup for robust fine-tuning.
\newblock {\em arXiv preprint arXiv:2203.03897}, 2022.

\bibitem{su2020predict}
Kun Su, Xiulong Liu, and Eli Shlizerman.
\newblock Predict \& cluster: Unsupervised skeleton based action recognition.
\newblock In {\em Proceedings of the IEEE/CVF Conference on Computer Vision and
  Pattern Recognition}, pages 9631--9640, 2020.

\bibitem{thoker2021skeleton}
Fida~Mohammad Thoker, Hazel Doughty, and Cees~GM Snoek.
\newblock Skeleton-contrastive 3d action representation learning.
\newblock In {\em Proceedings of the 29th ACM International Conference on
  Multimedia}, pages 1655--1663, 2021.

\bibitem{townsend2016pymanopt}
James Townsend, Niklas Koep, and Sebastian Weichwald.
\newblock Pymanopt: A python toolbox for optimization on manifolds using
  automatic differentiation.
\newblock {\em arXiv preprint arXiv:1603.03236}, 2016.

\bibitem{verma2019manifold}
Vikas Verma, Alex Lamb, Christopher Beckham, Amir Najafi, Ioannis Mitliagkas,
  David Lopez-Paz, and Yoshua Bengio.
\newblock Manifold mixup: Better representations by interpolating hidden
  states.
\newblock In {\em International Conference on Machine Learning}, pages
  6438--6447. PMLR, 2019.

\bibitem{pmlr-v162-wang22q}
Zekai Wang and Weiwei Liu.
\newblock Robustness verification for contrastive learning.
\newblock In Kamalika Chaudhuri, Stefanie Jegelka, Le Song, Csaba Szepesvari,
  Gang Niu, and Sivan Sabato, editors, {\em Proceedings of the 39th
  International Conference on Machine Learning}, volume 162 of {\em Proceedings
  of Machine Learning Research}, pages 22865--22883. PMLR, 17--23 Jul 2022.

\bibitem{wei2014skeleton}
Shih-En Wei, Nick~C Tang, Yen-Yu Lin, Ming-Fang Weng, and Hong-Yuan~Mark Liao.
\newblock Skeleton-augmented human action understanding by learning with
  progressively refined data.
\newblock In {\em Proceedings of the 1st ACM International Workshop on Human
  Centered Event Understanding from Multimedia}, pages 7--10, 2014.

\bibitem{weinzaepfel2021mimetics}
Philippe Weinzaepfel and Gr{\'e}gory Rogez.
\newblock Mimetics: Towards understanding human actions out of context.
\newblock {\em International Journal of Computer Vision}, 129(5):1675--1690,
  2021.

\bibitem{yan2018spatial}
Sijie Yan, Yuanjun Xiong, and Dahua Lin.
\newblock Spatial temporal graph convolutional networks for skeleton-based
  action recognition.
\newblock In {\em Thirty-second AAAI conference on artificial intelligence},
  2018.

\bibitem{yang2021skeleton}
Siyuan Yang, Jun Liu, Shijian Lu, Meng~Hwa Er, and Alex~C Kot.
\newblock Skeleton cloud colorization for unsupervised 3d action representation
  learning.
\newblock In {\em Proceedings of the IEEE/CVF International Conference on
  Computer Vision}, pages 13423--13433, 2021.

\bibitem{you2020graph}
Yuning You, Tianlong Chen, Yongduo Sui, Ting Chen, Zhangyang Wang, and Yang
  Shen.
\newblock Graph contrastive learning with augmentations.
\newblock {\em Advances in neural information processing systems},
  33:5812--5823, 2020.

\bibitem{yun2019cutmix}
Sangdoo Yun, Dongyoon Han, Seong~Joon Oh, Sanghyuk Chun, Junsuk Choe, and
  Youngjoon Yoo.
\newblock Cutmix: Regularization strategy to train strong classifiers with
  localizable features.
\newblock In {\em Proceedings of the IEEE/CVF international conference on
  computer vision}, pages 6023--6032, 2019.

\bibitem{zhang2017mixup}
Hongyi Zhang, Moustapha Cisse, Yann~N Dauphin, and David Lopez-Paz.
\newblock mixup: Beyond empirical risk minimization.
\newblock {\em arXiv preprint arXiv:1710.09412}, 2017.

\bibitem{zhang2022skeletal}
Haoyuan Zhang, Yonghong Hou, and Wenjing Zhang.
\newblock Skeletal twins: Unsupervised skeleton-based action representation
  learning.
\newblock In {\em 2022 IEEE International Conference on Multimedia and Expo
  (ICME)}, pages 1--6. IEEE, 2022.

\bibitem{zhang2016colorful}
Richard Zhang, Phillip Isola, and Alexei~A Efros.
\newblock Colorful image colorization.
\newblock In {\em European conference on computer vision}, pages 649--666.
  Springer, 2016.

\bibitem{zhang2022m}
Shaofeng Zhang, Meng Liu, Junchi Yan, Hengrui Zhang, Lingxiao Huang, Xiaokang
  Yang, and Pinyan Lu.
\newblock M-mix: Generating hard negatives via multi-sample mixing for
  contrastive learning.
\newblock In {\em Proceedings of the 28th ACM SIGKDD Conference on Knowledge
  Discovery and Data Mining}, pages 2461--2470, 2022.

\bibitem{zheng2018unsupervised}
Nenggan Zheng, Jun Wen, Risheng Liu, Liangqu Long, Jianhua Dai, and Zhefeng
  Gong.
\newblock Unsupervised representation learning with long-term dynamics for
  skeleton based action recognition.
\newblock In {\em Proceedings of the AAAI Conference on Artificial
  Intelligence}, volume~32, 2018.

\end{thebibliography}
}

\ifarxiv \clearpage \appendix
\label{sec:appendix}
\setlist[enumerate]{itemsep=0mm}
In this Supplementary material, we provide additional empirical studies, analyses, and details. We summarily list below the key sections.  

\begin{enumerate}
    \item Results on semi-supervised learning (Sec. \ref{sec:semisup})
    \item Additional ablation studies and analyses (Sec. \ref{sec:additional_ablations})
    \item Results with multi-modal ensembles (Sec.~\ref{sec:mm_results})
    \item HaLP for Graph Representation Learning (Sec. ~\ref{sec:halp_graph})
    \item HaLP with AimCLR (Sec.~\ref{sec:halp_otherpipelines})
    \item Implementation details (Sec.~\ref{sec:implementation_details})
\end{enumerate}

\section{Semi-supervised learning}
\label{sec:semisup}
In this experiment, we attach an MLP to the pre-trained backbone and finetune the MLP with labels on x\% of the training data. We follow the same setting as prior work and report results at 1\%, 5\%, 10\%, and 20\% data. 
In Table~\ref{tab:semi_sup_comparison}, we see that \ours{} outperforms the single modality baseline and prior state-of-the-art with comparable training showing that our representations can be adapted to the target task with little training data. HaLP with multi-modal training leads to results competitive with state-of-the-art. 
\begin{table*}

		\centering
\caption{Results on NTU-60 in semi-supervised learning setup. Following CMD \cite{mao2022cmd} we select a fraction of data and use that labelled data to fine-tune the network. We achieve consistent performance improvement over the other baselines for single-modality training. Using HaLP with CMD, gives improvements on NTU-60 x-sub and competetive performance on NTU-60 x-view. Results for AimCLR~\cite{Guo2022ContrastiveLF} were obtained using official models.}
		\begin{tabular}{lccccccccc}
			\toprule
			\multirow{3}{*}{Method} & \multicolumn{8}{c}{\textbf{NTU-60 }}\\
			\cmidrule(lr){2-9}
			& \multicolumn{4}{c}{\textbf{x-view}} & \multicolumn{4}{c}{\textbf{x-sub}} \\
			\cmidrule(lr){2-5} \cmidrule(lr){6-9}
			&   (1\%) &  (5\%) & (10\%) &  (20\%) & (1\%) &  (5\%) & (10\%) &  (20\%) \\
			\midrule
			\multicolumn{9}{l}{\textit{Additional training modalities or encoders}} \\
			ISC \cite{thoker2021skeleton} &  38.1   &  65.7   &72.5   &78.2   &35.7   &59.6   &65.9   &70.8 \\
			CrosSCLR-B \cite{li20213d} &     49.8   &  70.6   &77.0   &81.9   &48.6   &67.7   &72.4   &76.1 \\
			CMD ~\cite{li20213d}  & \textbf{53.0} &  \textbf{75.3} & 80.2  & 84.3  & 50.6 & 71.0   & 75.4  & 78.7 \\
			\textbf{\ours{}+CMD} & \textbf{53.0} & \textbf{75.3} & \textbf{80.4} & \textbf{84.6} & \textbf{52.6} & \textbf{71.4} & \textbf{76.0} & \textbf{79.2} \\
		    \hline
		    \multicolumn{9}{l}{\textit{Training using only joint}} \\
			LongT GAN \cite{zheng2018unsupervised}   &  - & - & - & - & 35.2 & - & 62.0 & - \\
			MS$^2$L \cite{Lin2020MS2LMS}  &  - & - & - & - & 33.1 & - & 65.1 & -  \\
			ASSL \cite{Si2020AdversarialSL} &  - & 63.6 & 69.8 & 74.7 & - & 57.3 & 64.3 & 68.0 \\
            AimCLR~\cite{Guo2022ContrastiveLF}$^\dagger$   &  47.2  & - & 74.6 & - & 45.7 & -  & 71.4 & -\\
			\ours{}-Baseline   &  45.4  & 69.0   & 75.2 & 81.0 & 42.6 & 64.3  & 70.0  & 74.7 \\
			\textbf{\ours{}}   &  \textbf{48.7}  & \textbf{71.5} & \textbf{77.1} & \textbf{82.4} & \textbf{46.6} & \textbf{66.9}  & \textbf{72.6}  & \textbf{76.1}\\
			\bottomrule
		\end{tabular}
\label{tab:semi_sup_comparison}
\end{table*}

\section{Ablation studies}
\label{sec:additional_ablations}
\begin{table}
\centering
\caption{
Different filtering approaches. We empirically evaluate the efficacy of various filtering approaches (defined in \cref{sec:filtering_approaches}). We see that while each of the approach leads to an improvement over the baseline, our rank filter outperforms the other proposed alternatives.
}
\begin{tabular}{lccc}
\toprule
Method & $G_{\text{filtered}}$ (/100) & NTU-60 x-sub \\
\midrule
Baseline & - & 78.0 \\
\hline
Rank filter & 91 & 79.7\\
variant-1 & 16 & 79.5\\
variant-2 & 100 & 79.6\\
\bottomrule
\end{tabular}

\vspace{-0.05in}
\label{tab:filtering_variants}
\end{table}

\subsection{Different filtering variants:} 
\label{sec:filtering_approaches}
As we discuss in the main paper, after the generation step we apply our \textbf{Rank Filter} (Main paper, ~\cref{eq:rank_filter} and Sec. 3.4) to filter out good positives for use with the loss. In this section, we experiment with alternative filtering approaches which can be used once the positives are generated. These are as follows: 
\\

\noindent\textbf{variant-1}: $\text{\zq}^\top\text{\zg} \geq \text{\zq}^\top\text{\zk}$ 
\\
Generated positives should be closer to the query features (\zq) than the key features (\zk).
\\

\noindent\textbf{variant-2}:
$\text{\zk}^\top\text{\zg} \geq P_{\text{sel}}^\top\text{\zk}$
\\
Generated positive should be closer to the key than the selected prototype.

In \cref{tab:filtering_variants}, we empirically compare the performance of these choices. We see that while all of our filters lead to improvements over the baseline, our proposed rank filter has the best performance. For these experiments, we generate 100 positives per anchor. We also report the number of retained positives after the filtering step. We see that since the rank filter has a stricter rule than variant-2, more positives get filtered out. The variant-1 is quite strict and retains 16 positives on average after the filtering step. These might be easier positives which might explain smaller improvements over the baseline.

\begin{table}[t]
\centering
\caption{
Effect of $\mu$ on linear evaluation performance. Recall that $\mu$ weighs the contribution of the HaLP loss compared to the standard contrastive loss. We see that increasing $\mu$ improves performance which saturates at higher values.}
\resizebox{0.6\linewidth}{!}{%
\begin{tabular}{lc}
        \toprule
        $\mu$ & NTU-60 x-sub \\
        \midrule
        0 (baseline) & 78.0 \\
        \hline
        0.5 &  78.8 \\
        1.0 & 79.7 \\
        1.5 & 79.9\\
        2.0 & 80.0 \\
        \bottomrule
        \end{tabular}
}

\label{tab:mu_ablation}
\end{table}

\subsection{Effect of changing $\mu$:} In \cref{tab:mu_ablation}, we vary the value of $\mu$ in the loss function (Main paper, ~\cref{eq:finalloss}) and observe its effect on the linear evaluation performance on the NTU60-xsub split. It is to be noted that as mentioned in the main paper, we use $\mu=0$ for the first 200 epochs. We empirically notice that this led to better performance (\cref{sec:when_to_use_mu}). First, we note that all of our approaches outperform the baseline (which has $\mu=0$ throughout). We observe that the performance steadily improves as $\mu$ is increased. For ease of experimentation, we set $\mu=1$ for all of our experiments. \footnote{Note that one could also set $\mu/\tau$ as a constant to be tuned but we refrain from doing so for making comparisons to the original MoCo loss easier.} 

\subsection{How many positives to generate:} This work proposes an efficient solution to hallucinate positives in the latent space. We can easily scale up the number of positives generated using our approach. In \cref{tab:num_positives}, we tabulate the performance by varying the number of positives. We see that using 100 positives has the best performance. As discussed in \cref{sec:method}, our approximation could generate positives that might not satisfy the constraint in \cref{eq:rank_filter}. Thus we pass all the generated positives through the \texttt{PosFilter()} before applying the loss. When using 100 positives, we observe that $\sim91\%$ positives are retained after the filtering step. We notice that other values of \# positives also outperform the baseline, which shows the efficacy of our approach.
\\

\begin{table}[t]
\centering
\caption{
Effect of $\#$ positives on linear evaluation performance. We empirically see that generating 100 positives shows the best performance.}
\resizebox{0.6\linewidth}{!}{%
\begin{tabular}{lc}
        \toprule
        \# Positives & NTU-60 x-sub \\
        \midrule
        0 (baseline) & 77.96 \\
        \hline
        1 &  79.59 \\
        5 & 79.63 \\
        10 & 79.50\\
        50 & 79.56 \\
        100 & \textbf{79.71} \\
        150 & 79.43 \\
        \bottomrule
        \end{tabular}
}

\label{tab:num_positives}
\end{table}

\subsection{Changing when the \ours{} loss is used:} 
\label{sec:when_to_use_mu}
For all experiments in the main paper, we set $\mu=0$ for the first 200 epochs and use a constant $\mu=1$ for the rest. In \cref{tab:mu_when_to_use_ablation}, we vary at which epoch $\mu$ is set to 1. We see that setting $\mu=1$ at 200 epochs leads to the best results. Note that the generation of positives is a function of the anchor and the prototypes. We hypothesize that using generated positives too early might be sub-optimal due to noisy early representations while using it too late might not give the model enough training signal to significantly improve the representations. 
\begin{table}[t]
\centering

\caption{In this expeirment, we vary when to start generating new positives. We notice that setting $\mu=1$ after 200 epochs gives the maximum improvement in performance. Note that all variants outperform the baseline.}
\resizebox{0.6\linewidth}{!}{\begin{tabular}{lc}
        \toprule
        $\mu=1$ at X epochs & NTU-60 x-sub \\
        \midrule
        50 & 79.6 \\
        100 & 79.6\\
        200 & \textbf{79.7} \\
        300 & 78.7\\
        400 & 78.3\\ 
        \hline
        $\mu=0$ (baseline) & 78.0 \\
        \bottomrule
        \end{tabular}
}

\label{tab:mu_when_to_use_ablation}
\end{table}

\subsection{Effect of topK}
Note that following the prior works ~\cite{thoker2021skeleton,mao2022cmd}, we use a memory queue of size 16384. Using the entire queue to obtain prototypes has two issues. First, this can increase the time to cluster. Secondly, note that the queue is obtained by appending the current keys in a first-in-first-out fashion. The use of very old (stale) keys can lead to sub-optimal representations since the model might have changed significantly. Thus we proposed to use only the most recent elements of the queue to obtain prototypes. In \cref{tab:queue_els}, we observe that using 256 most recent queue elements has the best performance.

        \begin{table}[t]
        \centering
\caption{
Effect of varying the number of Queue elements used to extract prototypes}
\resizebox{\linewidth}{!}{%
        \begin{tabular}{lccccc}
        \toprule
        \# Queue elements $\rightarrow$ & 256 & 512 & 1024 & 2048 & 3072\\
        \midrule
        NTU-60 x-sub & \textbf{79.7} & 79.5 & 79.4 & 79.5 & 79.7\\
        \bottomrule
        \end{tabular}
}

\label{tab:queue_els}
\end{table}

\subsection{Analyses of the generated positives}
\begin{figure}[!ht]
    \centering
    \begin{minipage}{\linewidth}
        \centering
        \includegraphics[width=\textwidth]{figs/files/sim_to_q.png}
        \caption{Mean similarity to query (\zq)}
        \label{fig:sim_query}
    \end{minipage}  
    \begin{minipage}{\linewidth}
        \centering
       \includegraphics[width=\textwidth]{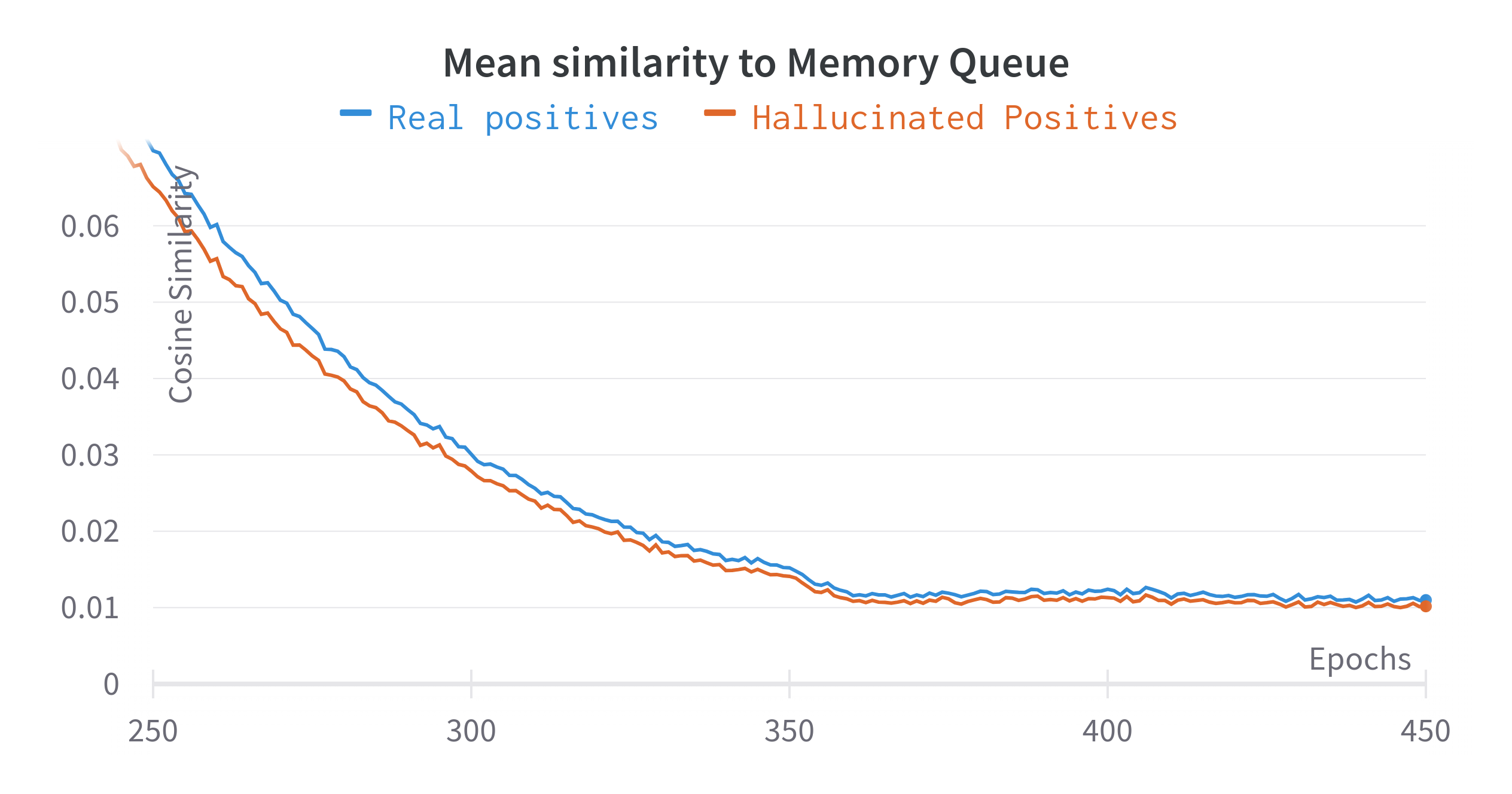}
    \caption{Mean similarity to Queue (negatives)}
    \label{fig:sim_queue}
    \end{minipage}
\end{figure}
\begin{figure}[tp]
        \centering
       \includegraphics[width=\linewidth]{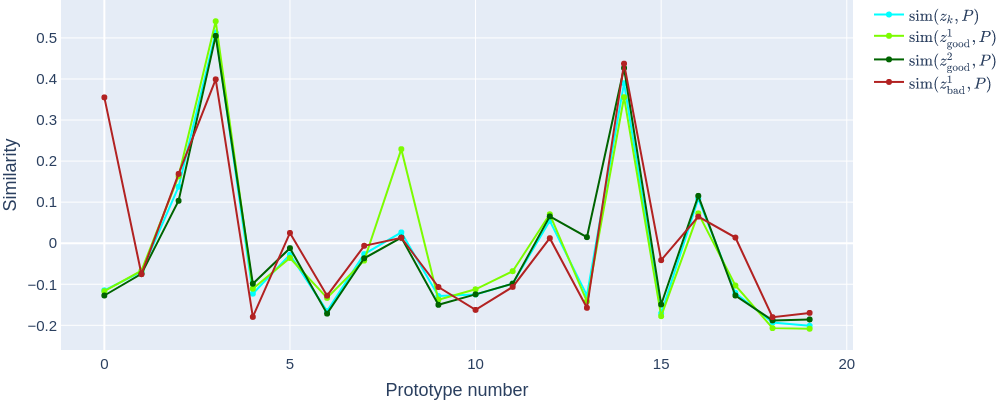}
    \caption{Similarity to prototype. We plot the similarity of the \zk and three generated positives to the prototypes. Two of the positives (green curves) share the same closest prototype as \zk (blue curve). These are the good positives which are used to compute the loss while the third positive (red curve) has a different closest prototype and is filtered out by the rank filter.}
    \label{fig:similarity_to_protype}
\end{figure}

\noindent\textbf{Evolution of similarities:} 
In \cref{fig:sim_query}, we plot the similarity of \zk and \zg to the query \zq for the last 200 epochs of training. In \cref{fig:sim_queue}, we plot the corresponding similarities to the queue (negatives). We see that the curves for real and hallucinated positives start with a larger gap but closely follow each other for the later part of training. We also notice that the mean similarities of positives are slightly higher for the real positives compared to the hallucinated ones. This verifies that our approach generates hard positives which lead to improved training. As expected, the anchor and positives are pulled closer together while the anchor and negatives are pushed farther away as training progresses. 

\noindent\textbf{Similarity of positives to prototypes:}
In \cref{fig:similarity_to_protype}, we plot the similarities of various positives to the prototypes.  First, $\text{sim}(z_k,P)$ (blue curve) represents the similarity of the original key to each of the prototypes. The key idea in our work is to generate positives that share the same closest prototype as the key \zk. We observe that the generated positives, ie positive 1 and positive 2 (green curves) satisfy the constraint above and add diversity to the generated positives while staying close to the original. We call these \emph{good positives}. Different from these, a \emph{bad positive} (red curve) significantly alters the similarities to each prototype and has a different closest prototype. This generated positive will get filtered out using our rank filter (Main paper, Sec. 3.4).

\subsection{HaLP w/o memory bank}
Most recent approaches (CMD, AimCLR, CrossSCLR, ISC) using contrastive learning use a memory queue due to its effectiveness. We can use HaLP without a memory queue by clustering the key features from the current batch or use the key features directly as cluster centers. We implement a Baseline w/o momentum using the former and evaluate the effectiveness of HaLP loss over it. We see that HaLP is beneficial (kNN NTU60-xsub: 58.9 $\rightarrow$ 64.2). Note that our approach also helps GraphCL ~(\cref{sec:halp_graph}) which does not use momentum queue either.

\section{Multi-modal ensemble}
\label{sec:mm_results}

Following prior work, we perform test-time ensemble of the 3 modalities : joint, motion, and bone. In \cref{tab:3s_comparison}, we see that our approach leads to improvements over the baseline. 

\begin{table}[t!]
	\centering
		\caption{Comparison with ensemble training. In this experiment, linear evaluation uses a late ensemble of various modalties. We see that \ours{}+CMD outperforms the multi-modal CMD~\cite{mao2022cmd} approach.}
		\resizebox{1.0\linewidth}{!}{
			\begin{tabular}{lccccc}
				\toprule
				\multirow{2}{*}{Method}  & \multicolumn{2}{c}{\textbf{NTU-60}}	& \multicolumn{2}{c}{\textbf{NTU-120}}	& \textbf{PKU-II} \\
				\cmidrule(lr){2-3} \cmidrule(lr){4-5} \cmidrule(lr){6-6}
				&  						 x-sub          & x-view  			& x-sub         	& x-set       & x-sub\\
				\midrule
				3s-CrosSCLR \cite{li20213d}			     		& 77.8           & 83.4   			& 67.9          & 66.7				& 21.2\\
				3s-AimCLR \cite{Guo2022ContrastiveLF}	         		& 78.9           & 83.8   			& 68.2          & 68.8 				& 39.5\\
				3s-CrosSCLR-B~\cite{li20213d}            		& 82.1  & 89.2	& 71.6 & 73.4 	& 51.0\\
				3s-CMD~\cite{mao2022cmd}            		& 84.1  & \textbf{90.9}	& 74.7 & 76.1 	& 52.6\\
				\textbf{3s-\ours{}+CMD}             		&  \textbf{85.1} & \textbf{91.0} & \textbf{75.5} &  \textbf{76.8}	& \textbf{53.4}\\
				\bottomrule
			\end{tabular}
		}
\label{tab:3s_comparison}
\end{table}

\section{HaLP for Graph Representation Learning}
\label{sec:halp_graph}
As a future work, we want to apply our work to behavior recognition in infants for early diagnosis of Autism Spectrum Disorder. We are particularly interested in skeleton-based action recognition since skeletons are grounded on people in videos, are less sensitive to scene and object biases and have minimal privacy concerns. These advantages of skeleton data make them beneficial for our eventual target task. Self-supervised learning on skeletons has received less attention due to difficulties in crafting geometrically consistent data augmentations. In~\cref{tab:graphcl_supplementary}, we apply our approach over GraphCL and see improvements over the baseline showing the general nature of our approach. 
\begin{table}[h]
        \centering
\caption{HaLP applied to Graph Representation Learning. We see that our approach of hallucinating positives also helps learn better encoders for general graphs. We use our plug-and-play module over GraphCL~\cite{you2020graph}}
\vspace{-0.15in}
\resizebox{\linewidth}{!}{%
        \begin{tabular}{lcccc}
        \toprule
        
         & NCI-1 & PROTEINS & DD & MUTAG\\
        \midrule
        GraphCL & $77.87 \pm 0.41$ & $74.39 \pm 0.45$ & $78.62 \pm 0.40 $& $86.80 \pm 1.3$\\
        +HaLP & $\mathbf{78.88 \pm 0.41}$ & $\mathbf{74.65 \pm 0.70}$ & $\mathbf{79.20 \pm 0.60}$ & $\mathbf{89.35 \pm 1.2}$ \\
        \bottomrule
        \end{tabular}
}
\label{tab:graphcl_supplementary}
\end{table}

\section{HaLP with AimCLR~\cite{Guo2022ContrastiveLF}}
\label{sec:halp_otherpipelines}
To show the plug-and-play nature of our approach, we add our module to AimCLR~\cite{Guo2022ContrastiveLF}, a recent Skeleton-SSL pipeline. We observe that adding HaLP is beneficial. The performance improves from 74.34 $\rightarrow$ \textbf{75.24} on NTU-60 x-sub showing the efficacy of our approach. 

\section{Implementation details}
\label{sec:implementation_details}
We use the same pre-training protocol as ISC~\cite{thoker2021skeleton} and CMD~\cite{mao2022cmd}. We use \texttt{Geomstats}~\cite{geomstats} for k-Means clustering on the hypersphere with tolerance of 1E-3 and initial step size of 1.0. We use $\lambda=0.8$ for all experiments. Wandb~\cite{wandb} was used for experiment tracking. Following are the key implementation details: 
\setlist[itemize]{itemsep=0mm}
    \begin{itemize}
        \item Epochs : 450 and 1000 epochs for NTU and PKU respectively. 
        \item Learning rate : 0.01, drop to 0.001 at 350 and 800 epochs for NTU and PKU respectively. 
        \item Memory queue size : 16384
        \item Batch size : 64
        \item Number of positives generated : 100
        \item Number of prototypes : 20 for NTU-60 and PKU and 40 for NTU-120. 
        \item We use $\mu=0$ from 0-200 epochs. For NTU-60 and PKU, $\mu$ is set to 1 at 200 epochs, while it is set to 2 for NTU-120. 
        \item Most recent elements used for clustering : 256
        \item Prototypes are updated every 5 steps.
    \end{itemize}

 \fi

\end{document}